\documentclass[11pt,a4paper]{article}
\usepackage[hyperref]{acl2020}
\usepackage{times}
\usepackage{latexsym}
\usepackage{booktabs}
\usepackage{multirow}
\usepackage{rotating}

\newcommand{\Sref}[1]{\S\ref{#1}}

\newcommand{\Fref}[1]{Figure~\ref{#1}}

\usepackage{microtype}
\usepackage{amsmath}
\usepackage{amssymb}

\aclfinalcopy 

\title{Demoting Racial Bias in Hate Speech Detection}

\author{Mengzhou Xia \quad Anjalie Field \quad Yulia Tsvetkov \\
  {Language Technologies Institute} \\ 
  {Carnegie Mellon University}  \\
  \texttt{\{mengzhox,anjalief,ytsvetko\}@cs.cmu.edu} 
  }
  
\date{}

\begin{document}
\maketitle
\begin{abstract}
In current hate speech datasets, there exists a high correlation between annotators' perceptions of toxicity and signals of African American English (AAE). This bias in annotated training data and the tendency of machine learning models to amplify it cause AAE text to often be mislabeled as abusive/offensive/hate speech with a high false positive rate by current hate speech classifiers. In this paper, we use adversarial training to mitigate this bias, introducing a hate speech classifier that learns to detect toxic sentences while demoting confounds corresponding to AAE texts. Experimental results on a hate speech dataset and an AAE dataset suggest that our method is able to substantially reduce the false positive rate for AAE text while only minimally affecting the performance of hate speech classification.
\end{abstract}

\section{Introduction}

The prevalence of toxic comments on social media and the mental toll on human moderators has generated much interest in automated systems for detecting hate speech and abusive language \cite{schmidt-wiegand-2017-survey, fortuna2018survey}, especially language that targets particular social groups \cite{silva2016analyzing, mondal2017measurement, mathew2019spread}. However, deploying these systems without careful consideration of social context can increase bias, marginalization, and exclusion \cite{bender2018data,waseem2016hateful}.

Most datasets currently used to train hate speech classifiers were collected through crowdsourced annotations \citep{davidson2017automated, founta2018large}, despite  the risk of annotator bias. \citet{waseem2016you} show that non-experts are more likely to label text as abusive than expert annotators, and  \citet{sap2019risk} show how lack of social context in annotation tasks further increases the risk of annotator bias, which can in turn lead to the marginalization of racial minorities. More specifically, annotators are more likely to label comments as abusive if they are written in African American English (AAE). These comments are assumed to be incorrectly labelled, as annotators do not mark them as abusive if they are properly primed with dialect and race information \citep{sap2019risk}. 

These biases in annotations are absorbed and amplified by automated classifiers. Classifiers trained on biased annotations are more likely to incorrectly label AAE text as abusive than non-AAE text: the false positive rate (FPR) is higher for AAE text, which risks further suppressing an already marginalized community. More formally, the disparity in FPR between groups is a violation of the Equality of Opportunity criterion, a commonly used metric of algorithmic fairness whose violation indicates discrimination \cite{hardt2016equality}. According to \citet{sap2019risk}, the false positive rate for hate speech/abusive language of the AAE dialect can reach as high as 46\%.

Thus, \citet{sap2019risk} reveal two related issues in the task of hate speech classification: the first is biases in existing annotations, and the second is model tendencies to absorb and even amplify biases from spurious correlations present in datasets \cite{zhao2017men, DBLP:journals/corr/abs-1809-07842}. While current datasets can be re-annotated, this process is time-consuming and expensive. Furthermore, even with perfect annotations, current hate speech detection models may still learn and amplify spurious correlations between AAE and abusive language \cite{zhao2017men, DBLP:journals/corr/abs-1809-07842}.






In this work, we present an adversarial approach to mitigating the risk of racial bias in hate speech classifiers, even when there might be annotation bias in the underlying training data. In \Sref{sec:model}, we describe our methodology in general terms, as it can be useful in any text classification task that seeks to predict a target attribute (here, toxicity) without basing predictions on a protected attribute (here, AAE).
Although we aim at preserving the utility of classification models, our primary goal is not to improve the raw performance over predicting the target attribute (hate speech detection), but rather to reduce the influence of the protected attribute.

In \Sref{sec:experiments} and \Sref{sec:results}, we evaluate how well our approach reduces the risk of racial bias in hate speech classification by measuring the FPR of AAE text, i.e., how often the model incorrectly labels AAE text as abusive.
We evaluate our methodology using two types of data: (1) a dataset inferred to be AAE using demographic information \cite{blodgett2016demographic}, and (2) datasets annotated for hate speech \cite{davidson2017automated, founta2018large} where we automatically infer AAE dialect and then demote indicators of AAE in corresponding hate speech classifiers. 
Overall, our approach decreases the dialectal information encoded by the hate speech model, leading to a 2.2--3.2 percent reduction in FPR for AAE text, without sacrificing the utility of hate speech classification. 

\begin{table*}[t]
\centering
\begin{tabular}{ll}
\toprule
Dataset & Example \\ \midrule
\citet{founta2018large} & I am hungry and I am dirty as hell {bruh}, need dat shower and {dem} calories \\ \midrule
\citet{blodgett2016demographic} & so much energy and time wasted hatin on someone when {alla} that {coulda} been   
\\ &  put towards {makin} yourself better.... a… https://t.co/awCg1nCt8t \\ \bottomrule
\end{tabular}

\caption{Example from \citet{founta2018large} and \citet{blodgett2016demographic} where the state-of-the-art model misclassifies innocuous tweets (inferred to be AAE) as abusive language. Our model correctly classifies these tweets as non-toxic.}
\label{tab:example}
\end{table*}
\section{Methodology}
\label{sec:model}

Our goal is to train a model that can predict a target attribute (abusive or not abusive language), but that does not base decisions off of confounds in data that result from protected attributes (e.g., AAE dialect). In order to achieve this, we use an adversarial objective, which discourages the model from encoding information about the protected attribute. Adversarial training is widely known for successfully adapting models to learn representations that are invariant to undesired attributes, such as demographics and topics, though they rarely disentangle attributes completely \cite{li-etal-2018-towards, elazar2018adversarial, kumar2019topics,lample2018multipleattribute,landeiro2019discovering}.


\paragraph{Model Architecture} Our demotion model consists of three parts: 1) 
An encoder $H$ that encodes the text into a high dimensional space; 2) A binary classifier $C$ that predicts the target attribute from the input text; 3) An adversary $D$ that predicts the protected attribute from the input text. We used a single-layer bidirectional LSTM encoder with an attention mechanism. Both classifiers are two-layer MLPs with a $\tanh$ activation function. 

\paragraph{Training Procedure} Each data point in our training set is a triplet $\{(x_i, y_i, z_i); i \in 1 \dots N \}$, where $x_i$ is the input text, $y_i$ is the label for the target attribute and $z_i$ is label of the protected attribute. The $(x_i, y_i)$ tuples are used to train the classifier $C$, and the $(x_i, z_i)$ tuple is used to train the adversary $D$.

We adapt a two-phase training procedure from \citet{kumar2019topics}. We use this procedure because \citet{kumar2019topics} show that their model is more effective than alternatives in a setting similar to ours, where the lexical indicators of the target and protected attributes are closely connected (e.g., words that are common in non-abusive AAE and are also common in abusive language datasets). In the first phase (pre-training), we use the standard supervised training objective to update encoder $H$ and classifier $C$:
\begin{gather}
    \min_{C, H} \sum_{i=1}^{N}  \mathcal{L}(C(H(x_i)), y_i)
    \label{eq:pretrain}
\end{gather}
After pre-training, the encoder should encode all relevant information that is useful for predicting the target attribute, including information predictive of the protected attribute.

In the second phase, starting from the best-performing checkpoint in the pre-training phase, we alternate training the adversary $D$ with \autoref{eq:tstep} and the other two models ($H$ and $C$) with \autoref{eq:cstep}:

\begin{align}
\begin{split}\label{eq:tstep}
      \min_D \frac{1}{N} \sum_{i=1}^{N} {}&  \mathcal{L}(D(H(x_i)), z_i)
\end{split}\\
\begin{split}\label{eq:cstep}
\min_{H, C}\frac{1}{N} \sum_{i=1}^{N} {}& \alpha \cdot \mathcal{L}(C(H(x_i)), y_i) + \\ 
& (1-\alpha) \cdot \mathcal{L}(D(H(x_i)), 0.5)  
\end{split}
\end{align}

Unlike \citet{kumar2019topics}, we introduce a hyper-parameter $\alpha$, which controls the balance between the two loss terms in \autoref{eq:cstep}. 
We find that $\alpha$ is crucial for correctly training the model (we detail this in \Sref{sec:experiments}). 

We first train the adversary to predict the protected attribute from the text representations outputted by the encoder.
We then train the encoder to ``fool'' the adversary by generating representations that will cause the adversary to output random guesses, rather than accurate predictions. At the same time, we train the classifier to predict the target attribute from the encoder output.



\section{Experiments}
\label{sec:experiments}

\subsection{Dataset}

To the best of our knowledge, there are no datasets that are annotated both for toxicity and for AAE dialect. Instead, we use two toxicity datasets and one English dialect dataset that are all from the same domain (Twitter):

 \paragraph{DWMW17 \cite{davidson2017automated}} A Twitter dataset that contains 25K tweets annotated as \textit{hate speech}, \textit{offensive}, or \textit{none}. The authors define hate speech as  language that is used to expresses hatred towards a targeted group or is intended to be derogatory, to humiliate,
or to insult the members of the group, and offensive language as language that contains offensive terms which are not necessarily inappropriate.
 \paragraph{FDCL18 \cite{founta2018large}} A Twitter dataset that contains 100K tweets annotated as \textit{hateful}, \textit{abusive}, \textit{spam} or \textit{none}. This labeling scheme was determined by conducting multiple rounds of crowdsourcing to understand how crowdworkers use different labels. Strongly impolite, rude, or hurtful language is considered abusive, and the definition of hate speech is the same as in DWMW17.
 \paragraph{BROD16 \cite{blodgett2016demographic}} A 20K sample out of a 1.15M English tweet corpus that is demographically associated with African American twitter users. Further analysis shows that the dataset contains significant linguistic features of African American English.

In order to obtain dialect labels for the DWMW17 and FDCL18, we use an off-the-shelf demographically-aligned ensemble model \citep{blodgett2016demographic} which learns a posterior topic distribution (topics corresponding to African American, Hispanic, White and Other) at a user, message, and word level.  \citet{blodgett2016demographic} generate a AAE-aligned corpus comprising tweets from users labelled with at least 80\% posterior probability as using AAE-associated terms. Similarly, following \citet{sap2019risk}, we assign AAE label to tweets with at least 80\% posterior probability of containing AAE-associated terms at the message level and consider all other tweets as Non-AAE. 

In order to obtain toxicity labels for the BROD16 dataset, we consider all tweets in this dataset to be non-toxic. This is a reasonable assumption since hate speech is relatively rare compared to the large amount of non-abusive language on social media \cite{founta2018large}.\footnote{We additionally did a simple check for abusive terms using a list of 20 hate speech words, randomly selected from \url{Hatebase.org}. We found that the percentage of sentences containing these words is much lower in AAE dataset ($\approx$ 2\%) than hate speech datasets ($\approx$ 20\%).}


\subsection{Training Parameters}
In the pre-training phase, we train the model until convergence and pick the best-performing checkpoint for fine-tuning. In the fine-tuning phase, we alternate training one single adversary and the classification model each for two epochs in one round and train for 10 rounds in total.

We additionally tuned the $\alpha$ parameter used to weight the loss terms in \autoref{eq:cstep} over validation sets. We found that the value of $\alpha$ is important for obtaining text representations containing less dialectal information. A large $\alpha$ easily leads to over-fitting and a drastic drop in validation accuracy for hate speech classification. However, a near zero $\alpha$ severely reduces both training and validation accuracy. We ultimately set $\alpha=0.05$.

We use the same architecture as \citet{sap2019risk} as a baseline model, which does not contain an adversarial objective. For both of this baseline model and our model, because of the goal of demoting the influence of AAE markers, we select the model with the lowest false positive rate on validation set. 
We train models on both DWMW17 and FDCL18 datasets, which we split into train/dev/test subsets following \citet{sap2019risk}.

\begin{table}[h]
    \centering
    \begin{tabular}{lcccc}
    \toprule
         \textbf{Dataset} &  \multicolumn{2}{c}{{\bf Accuracy}} & \multicolumn{2}{c}{{\bf F1}}\\
         \
         & base & ours & base & ours \\ \midrule
         DWMW17 & \textbf{91.90} & 90.68 & 75.15 & \textbf{76.05} \\
         FDCL18 & \textbf{81.18} & 80.27 & 66.15 & \textbf{66.80} \\ \bottomrule
    \end{tabular}
    \caption{Accuracy and F1 scores for detecting abusive language. F1 values are macro-averaged across all classification categories (e.g. hate, offensive, none for DWMW17). Our model achieves an accuracy and F1 on par with the baseline model.}
    \label{tab:accuracy}
\end{table}

\begin{table}[h]
    \centering
    \begin{tabular}{lcccc}
    \toprule
         &  \multicolumn{2}{c}{{\bf Offensive}} & \multicolumn{2}{c}{{\bf Hate}} \\
         & base & ours & base & ours \\
\midrule
         FDCL18-AAE & 20.94 &  \textbf{17.69}  & 3.23 & \textbf{2.60} \\
         BROD16 & 16.44 & \textbf{14.29} & 5.03 & \textbf{4.52} \\
         \bottomrule
         \end{tabular}
    \caption{False positive rates (FPR), indicating how often AAE text is incorrectly classified as hateful or abusive, when training with the FDCL18 dataset.  Our model consistently improves FPR for offensiveness, and performs slightly better than the baseline for hate speech detection.}
    \label{tab:false_pos_F}
\end{table}

\begin{table}[h]
    \centering
    \begin{tabular}{lcccc}
    \toprule
         &  \multicolumn{2}{c}{{\bf Offensive}} & \multicolumn{2}{c}{{\bf Hate}} \\
         & base & ours & base & ours \\ \midrule
         DWMW17-AAE & \textbf{38.27} & 42.59 & \textbf{0.70} & 2.06 \\
         BROD16 & \textbf{23.68} & 24.34 & \textbf{0.28} & 0.83     \\
         \bottomrule
         \end{tabular}
    \caption{False positive rates (FPR), indicating how often AAE text is incorrectly classified as hateful or offensive, when training with DWMW17 dataset. Our model fails to improve FPR over the baseline, since 97\% of AAE-labeled instances in the dataset are also labeled as toxic.
    }
    \label{tab:false_pos_D}
\end{table}


\section{Results and Analysis}
\label{sec:results}

\autoref{tab:accuracy} reports accuracy and F1 scores over the hate speech classification task. Despite the adversarial component in our model, which makes this task more difficult, our model achieves comparable accuracy as the baseline and even improves F1 score. Furthermore, the results of our baseline model are on par with those reported in \citet{sap2019risk}, which verifies the validity of our implementation.

Next, we assess how well our demotion model reduces the false positive rate in AAE text in two ways: (1) we use our trained hate speech detection model to classify text inferred as AAE in BROD16 dataset, in which we assume there is no hateful or offensive speech and (2) we use our trained hate speech detection model to classify the test partitions of the DWMW17 and FDCL18 datasets, which are annotated for hateful and offensive speech and for which we use an off-the-shelf model to infer dialect, as described in \Sref{sec:experiments}. Thus, for both evaluation criteria, we have or infer AAE labels and toxicity labels, and we can compute how often text inferred as AAE is misclassified as hateful, abusive, or offensive. 

Notably, \citet{sap2019risk} show that datasets that annotate text for hate speech without sufficient context---like DWMW17 and FDCL18---may suffer from inaccurate annotations, in that annotators are more likely to label non-abusive AAE text as abusive. However, despite the risk of inaccurate annotations, we can still use these datasets to evaluate racial bias in toxicity detection because of our focus on FPR. In particular, to analyze false positives, we need to analyze the classifier's predictions of the text as toxic, when annotators labeled it as non-toxic. \citet{sap2019risk} suggest that annotators over-estimate the toxicity in AAE text, meaning FPRs over the DWMW17 and FDCL18 test sets are actually lower-bounds, and the true FPR is could be even higher. Furthermore, if we assume that the DWMW17 and FDCL18 training sets contain biased annotations, as suggested by \citet{sap2019risk}, then a high FPR over the corresponding test sets suggests that the classification model amplifies bias in the training data, and labels non-toxic AAE text as toxic even when annotators did not.

\autoref{tab:false_pos_F} reports results for both evaluation criteria when we train the model on the FDCL18 data. In both cases, our model successfully reduces FPR. For abusive language detection in the FDCL18 test set, the reduction in FPR is $>3$; for hate speech detection, the FPR of our model is also reduced by $0.6$ compared to the baseline model. We can also observe a $2.2$ and $0.5$ reduction in FPR for abusive speech and hate speech respectively when evaluating on BROD16 data.

\autoref{tab:false_pos_D} reports results when we train the model on the DWMW17 dataset. Unlike \autoref{tab:false_pos_F}, unfortunately, our model fails to reduce the FPR rate for both offensive and hate speech of DWMW17 data. We also notice that our model trained with DWMW17 performs much worse than the model trained with FDCL18 data.

To understand the poor performance of our model when trained and evaluated on DWMW17 data, we investigated the data distribution in the test set and found that the vast majority of tweets labeled as AAE by the dialect classifier were also annotated as toxic (97\%). Thus, the subset of the data over which our model might improve FPR consists of merely $<3\%$ of the AAE portion of the test set (49 tweets). In comparison,  70.98\% of the tweets in the FDCL18 test set that were labeled as AAE were also annotated as toxic. Thus, we hypothesize that the performance of our model over the DWMW17 test set is not a representative estimate of how well our model reduces bias, because the improvable set in the DWMW17 is too small.

In \autoref{tab:example}, we provide two examples of tweets that the baseline classifier misclassifies abusive/offensive, but our model, correctly classifies as non-toxic. Both examples are drawn from a toxicity dataset and are classified as AAE by the dialectal prediction model.




\paragraph{Trade-off between FPR and Accuracy}
\begin{figure}[t]
  \includegraphics[width=\linewidth]{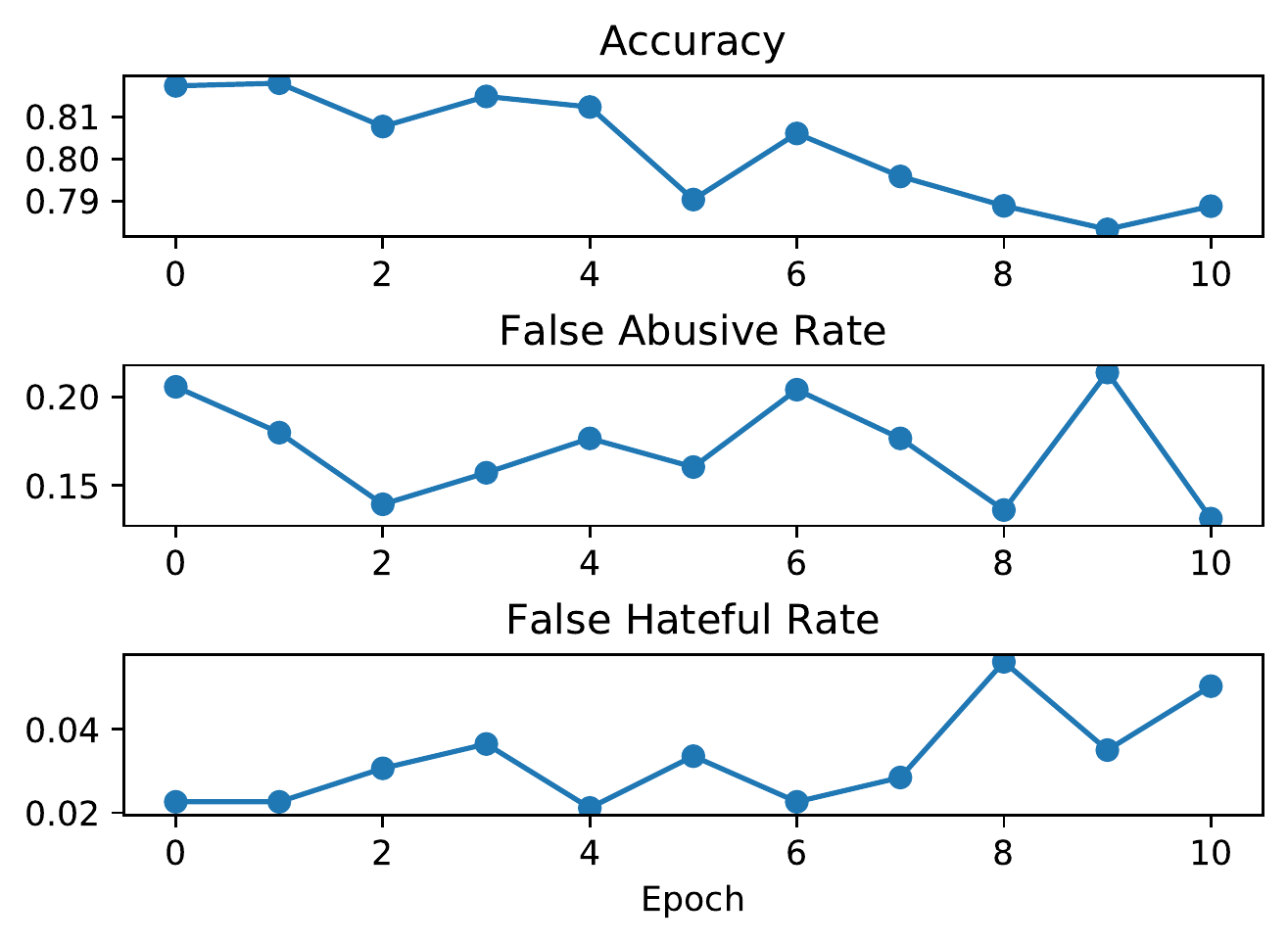}
  \caption{Accuracy of the entire development set of FDCL18 (top), and FPR rate for abusive (middle) and hate (bottom) speech detection for tweets inferred as AAE in the development set. X axis denotes the number of epochs. $0$th epoch is the best checkpoint for pre-training step, which is also the baseline model.}
  \label{fig:f2}
\end{figure}
In order to better understand model performance, we explored the accuracy and FPR of our model throughout the entire training process. We evaluate the best checkpoint of the pre-trained model ($0$\textsuperscript{th} epoch) and checkpoints of each epoch during adversarial training and show the results in \autoref{fig:f2}. While the baseline model ($0$\textsuperscript{th} epoch, before any adversarial training) achieves high accuracy, it also has a high FPR rate, particularly over abusive language. After adversarial training, the FPR rate decreases with only minor changes in accuracy. However, checkpoints with lower FPR rates also often have lower accuracy. While Tables \ref{tab:accuracy} and \ref{tab:false_pos_F} suggest that our model does achieve a balance between these metrics, \Fref{fig:f2} shows the difficulty of this task; that is, it is difficult to disentangle these attributes completely.

\begin{figure}[t]
  \includegraphics
  [width=\linewidth]{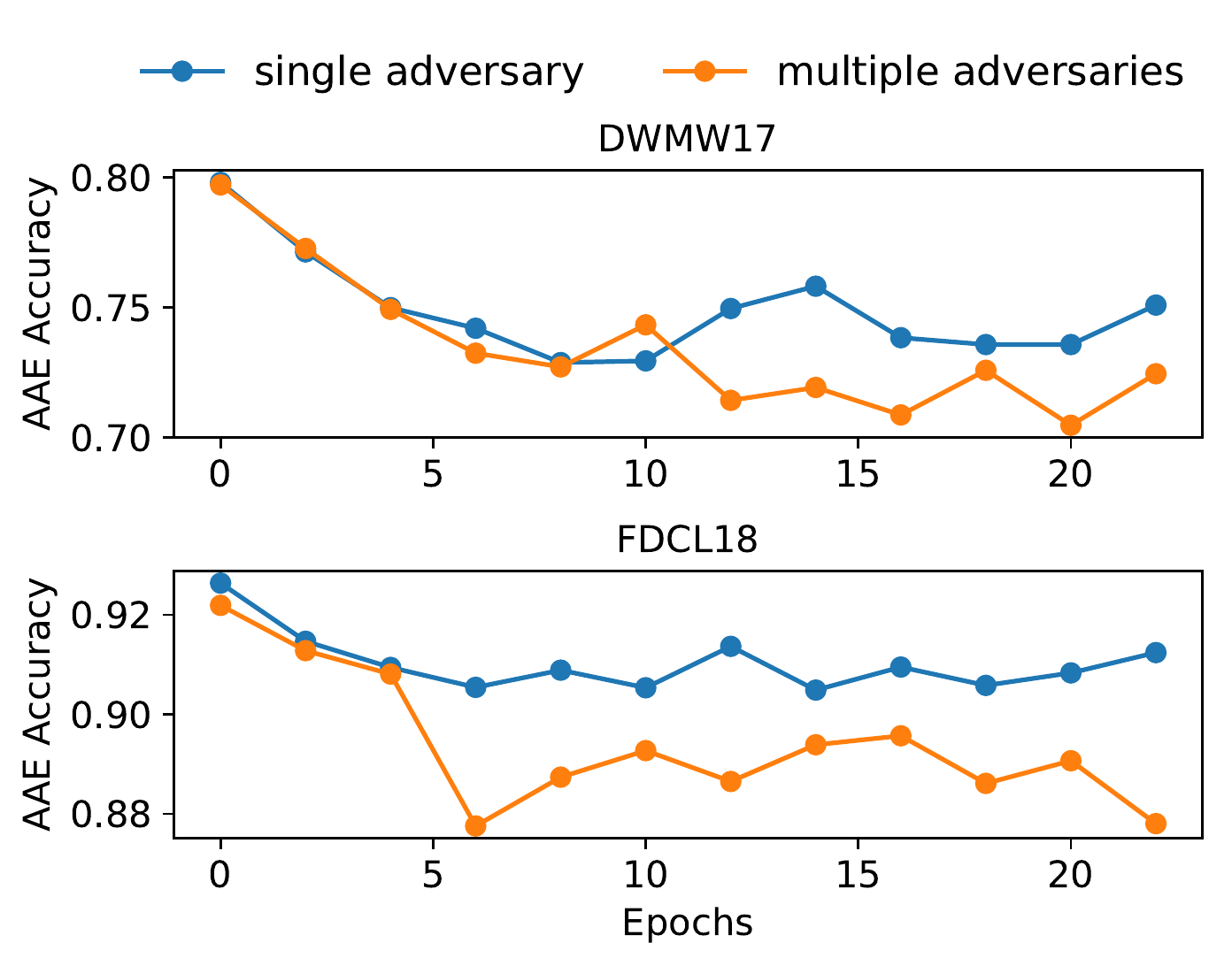}
  \caption{Validation accuracy on AAE prediction of the adversary in the whole training process. The green line denotes the training setting of one adversary and the orange line denotes the training setting of multiple adversaries.}
  \label{fig:f1}
\end{figure}

\paragraph{Eliminatation of protected attribute}
In \autoref{fig:f1}, we plot the validation accuracy of the adversary through the entire training process in order to verify that our model does learn a text representation at least partially free of dialectal information. Further, we compare using one adversary during training with using multiple adversaries \citep{kumar2019topics}. Through the course of training, the validation accuracy of AAE prediction decreases by about 6--10 and 2--5 points for both datasets, indicating that dialectal information is gradually removed from the encoded representation. However, after a certain training threshold (6 epochs for DWMW17 and 8 epochs for FDCL18), the accuracy of the classifier (not shown) also drops drastically, indicating that dialectal information cannot be completely eliminated from the text representation without also decreasing the accuracy of hate-speech classification. Multiple adversaries generally cause a greater decrease in AAE prediction than a single adversary, but do not necessarily lead to a lower FPR and a higher classification accuracy.  We attribute this to the difference in experimental setups: in our settings, we focus on one attribute to demote, whereas \citet{kumar2019topics} had to demote ten latent attributes and thus required multiple adversaries to stabilize the demotion model. Thus, unlike in  \cite{kumar2019topics}, our settings do not require multiple adversaries, and indeed, we do not see improvements from using multiple adversaries.

\section{Related Work}
Preventing neural models from absorbing or even amplifying unwanted artifacts  present in datasets is indispensable towards building machine learning systems without unwanted biases.  

One thread of work focuses on removing bias at the data level, through reducing annotator bias \cite{sap2019risk} and augmenting imbalanced datasets \cite{jurgens2017incorporating}. \citet{dixon2018measuring} propose an unsupervised method based on balancing the training set and employing a proposed measurement for mitigating unintended bias in text classification models. \citet{webster2018mind} present a gender-balanced dataset with ambiguous name-pair pronouns to provide diversity coverage for real-world data. In addition to annotator bias, sampling strategies also result in topic and author bias in datasets of abusive language detection, leading to decreased classification performance when testing in more realistic settings, necessitating the adoption of cross-domain evaluation for fairness  \cite{wiegand2019detection}.

A related thread of work on debiasing focuses at the model level \cite{zhao2019gender}. Adversarial training has been used to remove protected features from word embeddings \cite{xie2017controllable, zhang2018mitigating} and intermediate representations for both texts \cite{elazar2018adversarial, zhang2018mitigating} and images \cite{edwards2015censoring, wang2018adversarial}. Though previous works have documented that adversarial training fails to obliterate protected features, \citet{kumar2019topics} show that using multiple adversaries more effectively forces the removal. 

Along similar lines, multitask learning has been adopted for learning task-invariant representations. \citet{vaidya2019empirical} show that multitask training on a related task e.g., identity prediction, allows the model to shift focus to toxic-related elements in hate speech detection. 


\section{Conclusion}
In this work, we use adversarial training to demote a protected attribute (AAE dialect) when training a classifier to predict a target attribute (toxicity). While we focus on AAE dialect and toxicity, our methodology readily generalizes to other settings, such as reducing bias related to age, gender, or income-level in any other text classification task. Overall, our approach has the potential to improve fairness and reduce bias in NLP models.

\section{Acknowledgements}

We gratefully thank anonymous reviewers, Maarten Sap, and Dallas Card for their help with this work. The second author of this work is supported by the
NSF Graduate Research Fellowship Program under Grant No.~DGE1745016. Any opinions, findings, and conclusions or recommendations expressed in this material are those of the authors and do not necessarily reflect the views of the NSF. We also gratefully acknowledge Public Interest Technology University Network Grant No.~NVF-PITU-Carnegie Mellon University-Subgrant-009246-2019-10-01 for supporting this research. 

\bibliography{anthology,acl2020}
\bibliographystyle{acl_natbib}

\end{document}